# Addressing Biases in the Texts using an End-to-End Pipeline Approach

Shaina Raza[1](✉) , Syed Raza Bashir[2], Sneha[2], Urooj Qamar[3]

[1] University of Toronto, Toronto, ON, Canada
`shaina.raza@utoronto.ca`
[2] Toronto Metropolitan University, ON, Canada
`{syedraza.bashir, fnu.sneha}@torontomu.ca`
[3]Institute of Business & Information Technology, University of the Punjab, Lahore, Pakistan
`uroojqamar@ibitpu.edu.pk`

**Abstract.** The concept of fairness is gaining popularity in academia and industry. Social media is especially vulnerable to media biases and toxic language and comments. We propose a fair ML pipeline that takes a text as input and determines whether it contains biases and toxic content. Then, based on pre-trained word embeddings, it suggests a set of new words by substituting the biased words, the idea is to lessen the effects of those biases by replacing them with alternative words. We compare our approach to existing fairness models to determine its effectiveness. The results show that our proposed pipeline can detect, identify, and mitigate biases in social media data.

## 1 Introduction

Social media platforms allow users to interact with one another in a variety of ways, such as messaging, photo and video sharing apps and even allow users to leave comments and communicate with one another. This functionality is vulnerable to several internet crimes, including personal insults and threats, propaganda, fraud, and the advertisement of illegal goods and services. It is critical to identify and eliminate these toxic comments from social media that reflect biases.

The Conversation AI team, a joint venture between Jigsaw and Google develops technology to protect human voices during a conversation [1]. They are particularly interested in developing machine learning (ML) models that can detect toxicity in online conversations, with toxicity defined as anything biased, rude, disrespectful, offensive, or otherwise likely to cause someone to leave a discussion. This initiative has generated a substantial number of published words and competitions [2, 3].

In this paper, we propose a novel ML pipeline that ingests data and identifies toxic words early in the pre-processing stage; the identified words are then replaced with substitute words that retain the lexical meaning of the word but reduce or eliminate its effect. The main contribution of this work is to identify and mitigate biases during the pre-processing stage and avoid these biases replicate in the ML predictions. In this work, the term bias refers to any behavior, attitude, or expression that negatively affects a specific identity group, including actions that are hurtful, disrespectful, or disruptive. This definition is consistent with the bias definitions found in relevant literature [4–7]. The specific contribution of this work is as:

- We propose a fair ML pipeline that takes any data and detects if the biases exist, if existing then it mitigate those biases.
- We annotate the dataset with bias-bearing words, which are generally biased words used in toxic contexts to refer to specific identities (race, ethnicity, religion, gender), and are taken from various literature sources [8]; [9], and [10].
- We test each pipeline component individually to determine the method effectiveness, and we also quantify fairness (i.e., non-biased words in this context, for each sub-group based on identities- race, gender etc.).

## 2    Related Work

Fairness [11] is a multi-faceted concept that vary by culture and context. Bias mitigation or fairness methods are categorized into three broad types: (1) pre-processing; (2) in-processing; and (3) post-processing algorithms. The pre-processing algorithms [12] attempt to learn a new representation of data by removing the biases prior to algorithm training. In-processing algorithms influences the loss function during the model training to mitigate biases [13]. Post-processing algorithms [14] manipulate output predictions after training to reduce bias.

Several models have been proposed to address the issue of bias in ML algorithms and data. For example Fairness GAN [15] is a Generative Adversarial Network that learns to generate synthetic data samples to ensure demographic parity. Aequitas [16] is a toolkit for assessing and mitigating bias in predictive models. Themis-ML [17] is a library for creating fair ML models that utilizes algorithms for fairness-aware classification, regression, and clustering. Fairlearn [18] is another library for ML fairness built on top of the popular scikit-learn library. It provides metrics and algorithms for fairness evaluation and mitigation. Google's What-If Tool [19] is a visual interface for exploring ML models. It allows users to see the impact of changes to inputs and models on predictions and fairness metrics. AI Fairness 360 [20] is an open-source software toolkit that contains a comprehensive set of metrics, algorithms, and tutorials for detecting and mitigating bias in ML models. Other models such as Counterfactual Fairness [21], and Disentangled Representation Learning [22], also tackle the problem of bias mitigation in ML. It is important to note that while these models have shown promise in reducing bias, more research is needed to ensure the generalizability and effectiveness of these techniques. Each of the previous works is valuable and incremental, focusing on task fairness (pre / in / post-processing). Unlike previous works, we detect and mitigate many biases from text, and we build a pipeline to achieve fairness.

## 3  Proposed Methodology

We develop a fair ML pipeline (Figure 1) that accepts raw text, detects if the text is biased or not (detection task), then identifies the bias-bearing words in the text (recognition task), and finally substitutes those words with alternative words (mitigation task). We explain each phase next.

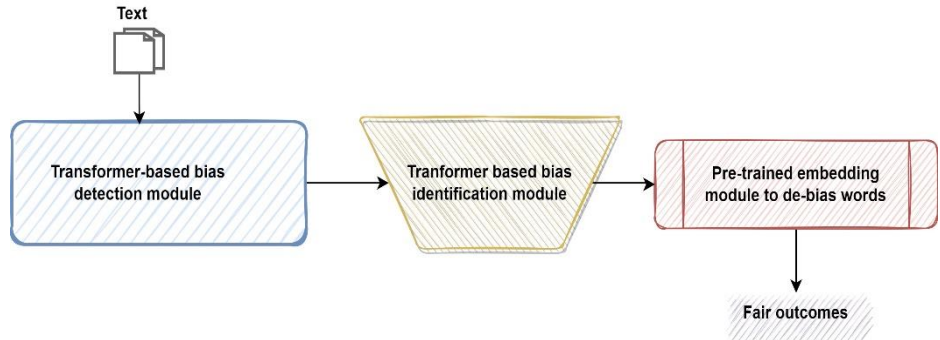

**Fig. 1.** Fair ML pipeline

*Bias detection:* The problem of bias detection involves identifying whether a given document, such as a news article or social media post, contains any biased language or perspectives. To address this problem, we have treated it as a multi-label classification task. Specifically, we have used a Transformer-based model called ELECTRA [23] and fine-tune the model for bias detection. We have used the labeled data provided by the Jigsaw Toxic Comment Classification [1] competition. This competition involved identifying whether comments on online forums were toxic or not. The dataset is also used in the competition to identify different types of language biases [1, 2] . By fine-tuning the ELECTRA model on this labeled data, we are able to adapt it to the specific task of bias detection. The output of the detection model is a sentence or text that has been labeled with one or more bias labels. These labels can indicate various types of biases, such as political bias, gender bias, or racial bias.

*Bias Identification:* The second step in the pipeline involves a module designed to identify biases within the dataset, which we refer to as the bias identification module. To create this module, we compiled a comprehensive list of biases that includes gender, race, religion, mental health, and disability. We also incorporated biases from sources such as [24], [8]; [9], and [10] to ensure that our list is comprehensive and up-to-date. Using this list of biases, we tag each comment in the dataset with relevant biases. Once the comments are tagged, we fine-tune the BERT model for named entity recognition (NER) to identify the biased words within the text. This fine-tuned model is then used to identify instances of bias in the comments, allowing us to analyze the extent and nature of biases present in the dataset.

In the bias identification task, certain categories of bias may be more easily or hardly detected depending on the nature of the biases and the dataset being analyzed. For example, some biases may be more explicit, while others may be more subtle or implicit. Similarly, some biases may be more prevalent in certain types of texts or domains, such as gender bias in job postings or racial bias in news articles. Based on our initial assessment of the data, we find that we are able to cover a range of topics, including online toxicity, hate speech, and misinformation.

*Bias mitigation:* After identifying the biased words in the text, our next step is to mitigate these biases by recommending alternative words that can be used in place of the biased words. We typically recommend between 5 to 10 substitute words per biased word, based on their similarity and appropriateness in the context of the text.

To generate these substitute words, we utilize publicly available pre-trained word embeddings, specifically Word2Vec [25], which operates in a 300-dimensional space. BERT can also be used to understand the contextual meaning of words and phrases in text data, and to fill in for the words as the substitute task. However, BERT can be computationally expensive and may require extensive training data to perform well. So we choose to work the Word2Vec in this paper.

Our method for identifying appropriate substitute words is based on semantic similarity and word analogy benchmarks [26]. By using this method, we aim to preserve the semantic information present in word embeddings while removing any biases that may be present in the text. The idea behind using Word2Vec here is to offer suitable substitutes that can help ensure a more equitable and inclusive representation of the target groups through words/ phrases.

## 4 Experimental setup

In this work, we use Google's Jigsaw Multilingual Toxic Comment Classification [1] dataset. It includes 223,549 annotated user comments collected from Wikipedia talk pages. These comments were annotated by human raters with six labels 'toxic', 'severe toxic, 'insult', 'threat', 'obscene', and 'identity hate'.

We use the F1-score (F1) for the accuracy, and a bias metric ROC-AUC (b-AUC) [2] to evaluate fairness. This bias metric combines several sub-metrics to balance overall performance. We also use the disparate impact ratio [27] to quantify fairness.

For our experiments, we utilized a standard set of hyperparameters to train our models, including a batch size of 16, a sequence length of 512, and 6 labels for the classification task. We trained the models for 10 epochs and optmize the learning rate in the range of 0.0001-0.001, the dropout rate in the range of 0.1-0.5, and the weight decay in the range of 0.0001-0.001. Our experiments were conducted on an NVIDIA P100 GPU with 32 GB RAM, and we implemented our models using TensorFlow. We fine-tuned our models using pre-trained weights from Huggingface.co. These settings ensured that our models were optimized for performance and accuracy.

## 5 Results

*Evaluation of bias detection task*: We evaluate our multi-label classifier with baseline methods: Logistic Regression with TFIDF (LG-TFIDF), LG with ELMO [28], BERT-base and DistillBERT.

Table 1. Performance of bias detection task. Bold means best performance.

| Model | b-AUC | F1 |
|---|---|---|
| LG-TFIDF | 0.547 | 0.585 |
| LG- ELMO | 0.684 | 0.625 |
| BERT-base | 0.692 | 0.687 |
| DistilBERT | 0.742 | 0.753 |
| Our model | **0.837** | **0.812** |

We observe in Table 1 that LG-TFIDF model has the lowest performance, achieving a b-AUC score of 0.547 and an F1-score of 0.585. The LG-ELMO model has an improved b-AUC score of 0.684 and F1 score of 0.625. The BERT-base model achieves a higher b-AUC score of 0.692, but its F1 score is comparatively lower at 0.687. Distilbert model achieves the b-AUC score of 0.742 and the F1 score of 0.753. Our model outperforms all other models, achieving the highest b-AUC score of 0.837 and the highest F1 score of 0.812. The significant improvement in the performance of our model suggests that it is effective in detecting bias in text data.

*Effectiveness of bias identification task*: We compare different configurations of NER: Spacy core web small (core-sm), core web medium (core-md), and core web large (core-lg) methods (that are based on RoBERTa [29]) against our NER.

Table 2. Performance of bias recognition task

| Model | AUC | F1 |
|---|---|---|
| Core-sm | 0.427 | 0.432 |
| Core-md | 0.532 | 0.524 |
| Core-lg | 0.643 | 0.637 |
| Our model | **0.832** | **0.828** |

Based on the performance metrics reported in Table 2, it is clear that our model outperformed the three baseline models (Core-sm, Core-md, Core-lg) by a significant margin in both AUC and F1 score. Our model achieved an AUC of 0.832 and F1 score of 0.828, while the best-performing baseline (Core-lg) achieved an AUC of 0.643 and F1 score of 0.637. This indicates that our model is fine-tuned properly on the biased labels and is more effective in recognizing bias in the dataset than the baseline models. It is also worth noting that the performance of the baseline models improved as the model size increased, with Core-lg performing better than Core-md and Core-sm. This also suggests that the size of the model can have a significant impact on its performance.

*Overall Performance comparison*: To evaluate the pipeline as a whole, we use the adversarial debiasing (AD) [13] and meta-fair (MF) classifier [30] methods as the baselines. AD is a fairness method that addresses fairness during the data pre-processing time and MF is an in-processing method that addressed biases during the optimization phase.

In this experiment, we provide the labeled data to each method. First, we use our detection module to find if a text is biased or not, and then use each method's debiasing technique to introduce fairness in the data. The new data that is produced in the transformed data. These methods calculate fairness based on the ratio of fair out-comes (non-biased words) for each sub-group (e.g., gender, and identities). For example, these methods see how many biased or unbiased words are associated with each identity group, and then remove the biases for the subgroup that is more prone to negative outcomes.

We consider the sub-groups based on gender and race as the use cases. For gender, we consider the privileged class to be "male," while the unprivileged class is "female". For race, we consider "Asians" and "American-Africans" to be unprivileged, and "white" to be privileged. These groups are chosen based on an initial analysis of the data. We use the disparate impact ratio evaluation metric to quantify fairness. A good range of DI ratio is between 0.8 and 1.25 [27] with scores lower than 0.8 showing favorable outcomes for privileged sub-group and values above 1.25 favoring the unprivileged class. The results are shown in Figure 2.

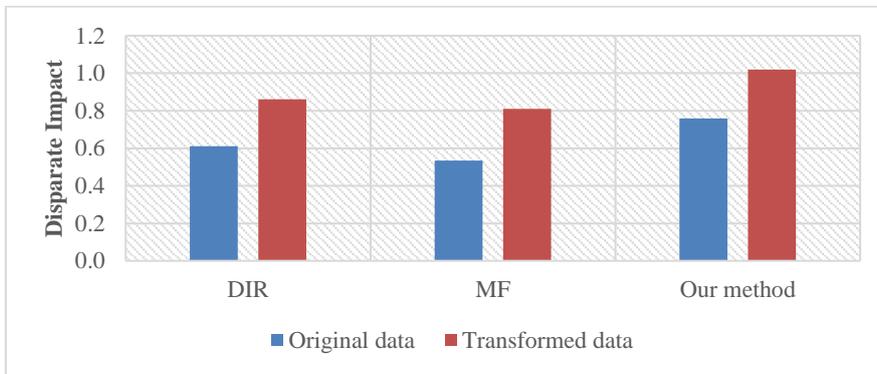

**Fig. 2.** Disparate impact scores to quantify fairness using different methods.

The results in Figure 2 show that the DI score in the original data is lower than 0.8, which means biased outcomes toward unprivileged identities. Before mitigation, the DI score is the same for all methods, since it is calculated based on original data. The DI score after fairness methods is above 0.8 for all methods. Our approach gives us a DI score close to 1, which shows that we achieve a balance between unprivileged and privileged groups. Other methods also get fairer on transformed data, but they seem to skewed toward privileged groups (score close to 0.8).

## 6 Discussion

The main implications of the proposed model are in applications where bias in textual data can have significant real-world impacts, such as in hiring and admissions decisions, financial lending, and predictive policing. Prior work [15] [16] [17] [18] [19] [20] [31] [13] [21] [32] in this area has also explored various techniques for detecting and mitigating text biases. However, the proposed method has the advantage of being a scalable and easy-to-implement solution that does not require additional annotation or training data. There are some limitations of this study, which are also suggestions for future work. First, the current work assumes that the biased words can be easily identified and replaced with alternative words, which may not always be the case. We need to consider the epistemology and tone in the biased language as well. The method also relies on pre-trained embeddings, which may contain biases and affect the quality of the mitigation. Further, the effectiveness of the method may vary across different domains and languages, which needs to be further investigated.

## 7    Conclusion

The proposed fair ML pipeline for detecting and mitigating text biases is an important step towards developing more equitable and just AI models. However, there is still much to learn about applying and interpreting fairness in our study. We plan to further evaluate the pipeline using other evaluation metrics and extend it to other domains such as health and life sciences. The bias mitigation task will also be evaluated to enhance the effectiveness of the pipeline. Additionally, we will explore other datasets to see how the pipeline works in other contexts. Through continuous evaluation and refinement, we aim to develop more sophisticated and effective fairness approach.